# Metaheuristics for the operating theater planning and scheduling: A systematic review

Amirhossein Moosavi[*], Onur Ozturk

*Telfer School of Management, University of Ottawa, 55 Laurier Avenue East, Ottawa, Ontario K1N 6N5, Canada.*

***Corresponding author***: Amirhossein Moosavi

Email address: amir.moosavi@uottawa.ca

ORCID (Amirhossein Moosavi): https://orcid.org/0000-0003-0456-8130

# Metaheuristics for the operating theater planning and scheduling: A systematic review

## Abstract

Healthcare expenses represent a large share of most developing countries' GDP. Operational theatres make up the majority of these costs in hospitals. There are found a vast number of papers studying the problem of operating theater planning and scheduling. Different variants of this problem are generally recognized to be NP-complete; thus, several solution approaches have been utilized in the literature to confront with these complicated problems. The lack of a thorough review of the main characteristics of solution approaches is tangible in the literature (reviewing them separately and with regards to the characteristics of studied problems), which can provide pragmatic guidelines for practitioners and future research projects. This paper aims to address this issue. Since different types of solution approaches usually have different characteristics, this paper focuses only on metaheuristic algorithms. Through both automatic and manual search methods, we have selected and reviewed 28 papers with respect to their main problem and solution approach features. Finally, some directions are introduced for future research.

*Keywords*: Operating Theater: Operating Room; Planning; Scheduling: Metaheuristic.

## 1. Introduction

Health care costs account for a big portion of GDP in most developed countries, e.g., 16.9% of the US's GDP and 10.7% of Canada's GDP in 2018 (OECD, 2020); The majority of these expenditures were spent in hospitals (National Health Expenditure Database, 2019). It is also known that more than 50% of hospitals' costs are incurred by Operating Theaters (OTs) (Denton et al., 2010) due to their expensive resources (Latorre-Núñez et al., 2016), usually high volume of demand (Moosavi and Ebrahimnejad, 2018) and complex resource allocation decisions. For example, Childers and Maggard-Gibbons (2018) reported that a functioning Operating Room (OR) in an OT costs $2,160 to $2,220 per hour on average. With an aging population, the need and cost for surgical services are likely to increase. All this underlines the importance of ensuring the efficient use of resources in OTs.

OTs are facilities within hospitals providing surgical services. An OT includes upstream units (e.g., a stepdown unit), Operating Rooms (ORs), and downstream units (e.g., a post-anesthesia care unit). A typical schema for this theater has been illustrated in Figure 1. As shown in this figure, a general physician first assesses the health status of an elective patient to see whether further assessment by a surgeon is needed. Elective patients referred to the surgeon assessment will be added to the waiting list if a surgery is required. Otherwise, the patient will be returned back to the general physician. For all elective patients in the waiting time, the date and time of surgery will be determined, i.e., the scheduling process. In the meantime, admitted emergency arrivals will go through the scheduling process as well. In the next step, patients (both elective and emergency) will be served by the OT: (1) they may be hospitalized in upstream units, such as stepdown unit, (2) they will be operated in ORs, and (3) they may be hospitalized in downstream units, such care unit or ward, or discharged.

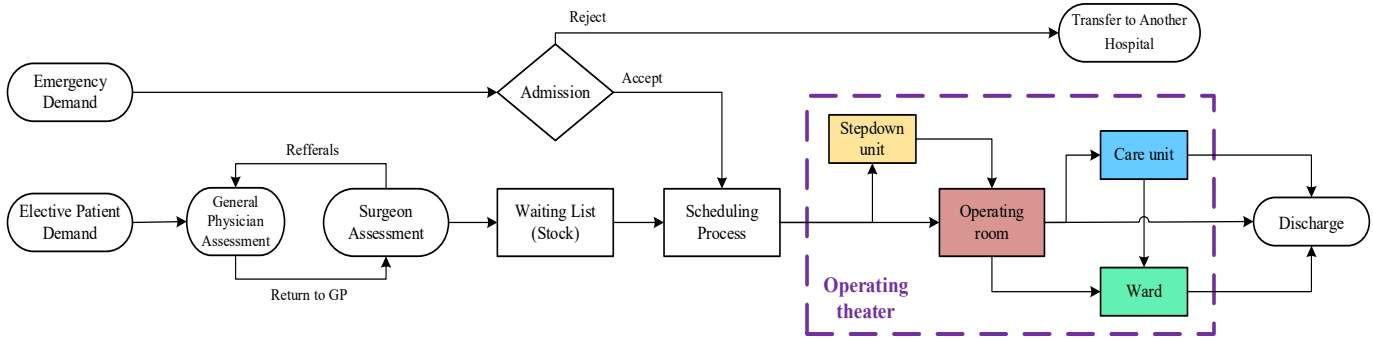

Figure 1. A typical schema for the flow of patients in a hospital (Moosavi and Ebrahimnejad, 2020)

According to Guido and Conforti (2017), OT Planning and Scheduling (OTPS) systems could be grouped as follows: open scheduling, block scheduling, and modified-block scheduling. The open scheduling system assigns the available time to the first surgeon who requests it. In the block scheduling system, each OR-block is allocated to a specialty; this scheduling system does not allow that two dissimilar specialties operate within an OR-block. Note that a block here refers to an interval of time. The modified-block scheduling system integrates previous scheduling systems in order to make use of their advantages.

The OTPS problems generally include three hierarchical decision levels (Zhu et al., 2018): strategic, tactical, and operational. The management and experts specify the time that each specialty can operate in ORs at the strategic level (case mix). The allocation of available ORs to each specialty is determined at the tactical level (master surgical scheduling – only for block scheduling system). Finally, the allocation and sequencing of surgical cases take place at the operational level (advance and allocation scheduling problems, respectively). Providing further information is not in the scope of our paper. For detailed information, interested readers are referred to Moosavi and Ebrahimnejad (2018). Note that we will use the definitions of these decision levels (or types of problems) to conduct the literature review.

Aringhieri et al. (2013) state that OTPS problems are usually highly constrained, and therefore, computationally complex (NP-complete). Previous studies have proposed different types of solutions approaches in order to solve large-scale instances of such problems. These solution approaches include mathematical programming, heuristics, model-based heuristics, metaheuristics, simulation, and others. As an important element of studies in OTPS problems, having familiarity with the state-of-the-art solution approaches can be of great help for current practitioners and future research.

The lack of a through review of the main characteristics of solution approaches is tangible in the literature (reviewing them separately and with regards to the characteristics of studied problems), which can provide pragmatic guidelines for current real-world applications and future research projects. In order to prove this argument, we searched the literature to find review papers of OTPS. Through this search, we found 28 review papers published between 1976 and 2019. For the sake of brevity, further details regarding the methodology of search and a list of all found review papers stored on the web at the address https://bitbucket.org/Pro_Data/slr. A brief summary for those review papers published after 2015, including Zhu et al. (2019), Gür and Eren (2018), Samudra et al. (2016), Van Riet and Demeulemeester (2015), and Wilton and Peter (2015), have been provided in Table 1.

To the best of our knowledge, none of the aforementioned review papers has evaluated the OTPS problems with an emphasis on the solution approach. For example, Zhu et al. (2019) extended the previous review papers by considering more analysis dimensions and adding more references. They categorized the literature into six groups, where one group is devoted to the solution approaches. In the section of solution approaches, the authors mostly discussed the features of previous studies rather than explaining their solution approaches in detail. Let us look at an example from Zhu et al. (2019):

"Fei et al. (2006) solve a weekly OR planning problem of allocating patients to blocks using a column-generation-based heuristic and address the following daily OR scheduling problem of determining the sequence of the patients with a hybrid genetic algorithm".

Including more details regarding the solution approaches was possible and beneficial. This deficiency is reasonable for their review paper because this 50-page paper covers different aspects of the OTPS problem. Furthermore, almost all the aforementioned papers are not Systematic Literature Reviews (SLRs), e.g., Zhu et al. (2019) and Wilton and Peter (2015) did not indicate what keywords or inclusion/exclusion criteria were used to find relevant references. This is while SLRs can synthesize existing research: (1) fairly (without bias), (2) rigorously (according to a defined procedure or protocol), and (3) reproducibly (ensuring that the review procedure is visible to and auditable by other researchers). Due to all reasons mentioned earlier, this paper aims to provide an SLR focusing on the solution approaches applied to OTPS problems, particularly metaheuristic algorithms. The current research tries to address two following research questions:

- What types of OTPS problems have been solved by metaheuristics?
- What are the most important features of metaheuristics applied to OTPS problems?

Table 1. A summary of five review papers published after 2015

| Literature review | Review period | Number of papers | Databases | Search query | Objectives |
|---|---|---|---|---|---|
| Zhu et al. (2019) | 2010 - 2019 | 236 | PubMed, Web of Science, IEEExplore, Springer and Inspec | Not provided | Extension of previous review studies by considering more analysis dimensions as well as referring more references |
| Gür and Eren (2018) | 2000 - 2018 | 170 | Emerald, Science Direct, JSTOR, Springer, Taylor and Francis, and Google Scholar | Not provided | Analyzing and general information about previous studies in the OTPS |
| Samudra et al. (2016) | 2000 - 2014 | 283 | PubMed and Web of Science | Surgery, case, operating, room, theat*, scheduling, planning and sequencing | Classify recent OR planning and scheduling, look for evolutions over time and identify the common pitfalls |
| Van Riet and Demeulemeester (2015) | 1990 - 2014 | 106 | Web of Science | Emergent surgery planning/ scheduling, emergency theater, semiurgent surgery planning, urgent surgery planning/scheduling, nonelective patient scheduling, emergency operating room, dedicated operating room capacity and operating room capacity emergency | Classify the literature that has included non-electives |
| Wilton and Peter (2015) | Not provided | 35 | Not provided | Not provided | Emphasize the efficient use of resources in OTs and discuss methods that can be applied to add value to health organization through optimal OTPS management |

The remainder of this paper is organized as follows. In Section 2, the protocol of this SLR will be discussed in detail. Then, we will evaluate the features of OTPS problems solved by metaheuristics in Section 3 in order to answer the first research question. For the second research question, features of the applied metaheuristic algorithms will be discussed in detail in Section 4. Finally, Section 5 presents conclusions, limitations of this investigation, and future research directions.

## 2. Research protocol

A well-devised search strategy is a fundamental part of an SLR. Weak search strategies will result in a prejudiced or inadequate list of references. As a starting point, we have gone through the previous review papers and found what keywords have been used frequently. Using these keywords, we can construct a search query. Ideally, I would like to have a search query, which does not provide me a sheer/ a few numbers of results. In other words, the number of results must be adequately large to have an impartial SLR; at the same time, it should be small enough such that conducting the SLR would be feasible.

After determining the search query, we have applied the automatic search method (a search through online databases via a search query) to find relevant papers. Then, we have applied a manual search method, i.e., forward and backward snowballing, to find potential further papers. Finally, using some inclusion and exclusion criteria, and some quality assessment criteria, we have eliminated irrelevant papers. A detailed description of different steps of our SLR has been provided in a protocol stored on the web at the address https://bitbucket.org/Pro_Data/slr. Nevertheless, for the sake of completeness, a brief explanation has been provided for each step of our SLR in the following subsections.

### 2.1. Source databases

The decision to include a database depends on the topic of the problem. The exclusion of an important database could lead to a biased and incomplete SLR. Databases included in this SLR are Scopus, Web of Science, PubMed and IEEEXplore. Scopus and Web of Science are two general and multipurpose databases. PubMed is the biggest and most used database in medical science studies. Finally, IEEEXplore is a large engineering database where I can possibly find relevant papers.

### 2.2. Search query and databases

A good search query plays an important role in having high precision and recall rates. The search query used in the current SLR is as follows:

(("*operating theat\**" **OR** "*operating room*") **NEAR/2** (*planning* **OR** *schedul\**)) **AND** (*optimization* **OR** "*mathematical programming*" **OR** "*mathematical model\**" **OR** \**heuristic\**))

Because search operators of different databases vary slightly, therefore, the above search query must be adjusted for each of them. For example, the proximity operator of "NEAR" is merely applicable in Web of Science and IEEEXplore. This proximity operator for Scopus, however, is "W". PubMed does not support proximity operators; consequently, we have replaced this operator with "AND" for PubMed.

### 2.3. Exclusion and inclusion criteria

To make an SLR more purposeful and viable, suitable exclusion criteria can be applied to eliminate irrelevant papers (narrow down the focus of research). The exclusion criteria of this systematic literature review are as follows:

- Exclude published books, chapters, reports, abstracts, and theses

- Exclude papers published in languages other than English

- Exclude working and under review documents by the date of the submission of this review paper

- Exclude documents without having access to the entire content

- Exclude non-peer reviewed documents (to make sure that all documents meet a certain level of academic quality)

- Exclude documents published before 2015

On the other hand, because the automated search method might not have found all relevant papers, inclusion criteria can be applied to make sure that most relevant resources are included (preventing biased results). The only inclusion criterion of this paper is the application of the snowballing method (a manual search method). The snowballing method uses (an) exemplar paper(s). Then, it tries to find undiscovered relevant papers through references of the exemplar paper(s) (backward approach) and tracking the citation of that specific paper(s) (forward approach). We have used Zhu et al. (2019) and Gür and Eren (2018) as exemplars due to two main reasons: (1) these are two of the most up-to-date review papers for OT planning and scheduling problems, and (2) these papers cover 236 and 170 papers, respectively.

To eliminate duplicates (papers found by different databases during the automated search method), we have used Covidence.org. In addition, we have only gone through the title, abstract, and keywords of the papers to be able in order to apply both exclusion and inclusion criteria.

### 2.4. Quality assessment criteria

Quality assessment criteria are generally used to evaluate the quality of the retrieved papers. These criteria are useful to make an SLR more practical and focus on a specific part of the intended problem. For this study, we have defined two main quality assessment criteria as follows:

- Is the problem under consideration explained explicitly with regard to its assumptions, objective functions, and constraints?
- Is the solution approach evaluated by benchmark, randomly generated, real test examples?

To apply these quality assessment criteria, we have screened the text of each paper. If a paper does not satisfy each of these criteria, it will be eliminated.

### 2.5. Search summary

To summarize the search procedure of this SLR, Figure 2 illustrates a PRISMA flow chart. According to this figure, 294 papers have been found in the identification step (through both automated and manual search methods). Using Covidence.org, 119 duplicates have been recognized and eliminated. In the next step, 141 irrelevant papers have been found through the application of exclusion criteria (105 through screening the title, abstract and keywords, and 36 through screening the full text). On the other hand, the manual search method has found 8 additional relevant papers. It means that 42 papers have been passed to the next step. Finally, all papers were fully assessed through the application of quality assessment criteria, and 14 papers have been eliminated. In total, 28 papers have been entered into the data extraction step.

Two recent review papers, i.e., Zhu et al. (2019) and Gür and Eren (2018), reported only 18 more applications of metaheuristics from 2000 until 2014. Only eight of these papers meet both our exclusion criteria (except the last one) and quality assessment criteria. Thus, these criteria are not limiting the number of papers included in this review. In addition, we prefer to merely include those 28 papers published between 2015 and 2020; because these papers could be better a representative of recent and more computationally complex.

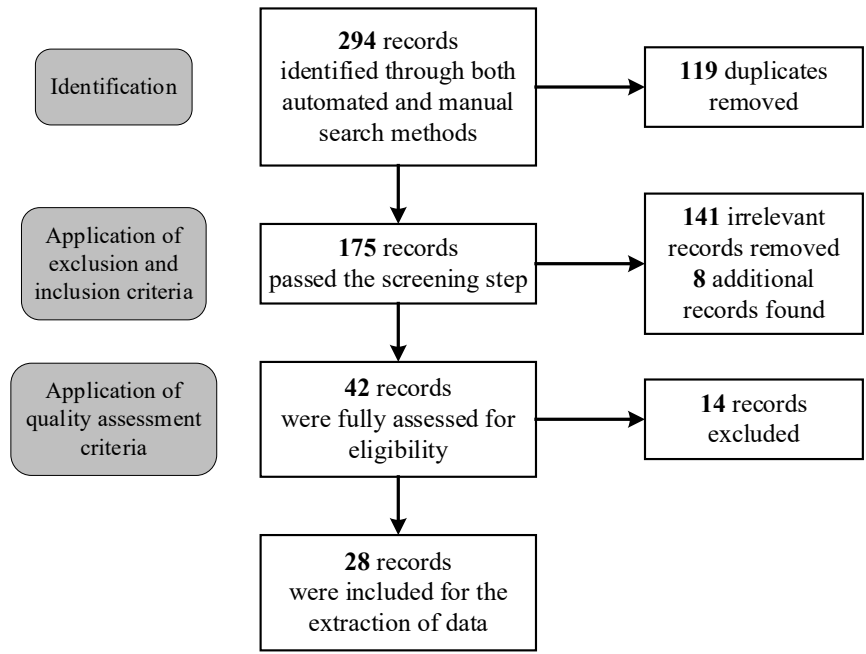

Figure 2. The PRISMA flow chart of the current SLR

## 3. Operating theater planning and scheduling

In this section, the 28 selected papers have been reviewed in terms of their problem features, which let us answer the first research question of this SLR. First of all, we have categorized these papers based on the year and type of their publication in Figure 3. As shown in Figure 3, 11 papers have been published in 2018 (the greatest frequency), while no paper has been published in 2020 – until the publication date – (the least frequency). Looking at these papers, we also realized that the majority of publications were journal papers (21 publications); whereas only seven of them were conference papers.

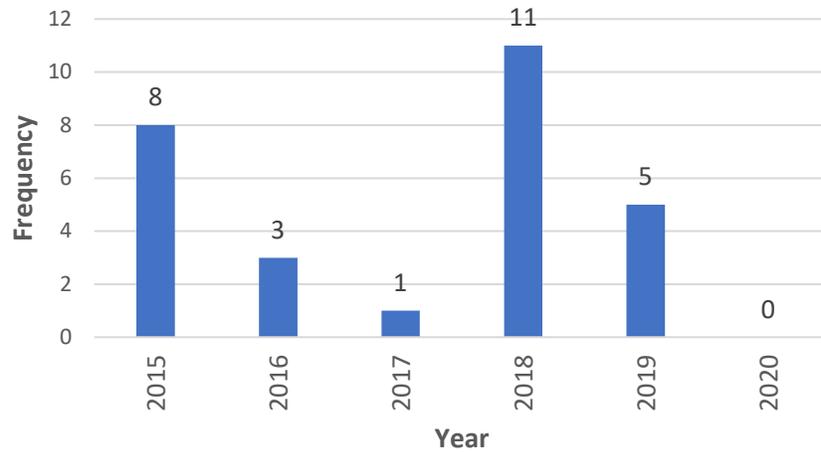

Figure 3. Number of publications categorized based on the year and type

The application of metaheuristics depends on the type of OTPS problem. Although the focus of this review is on the investigation of solution approaches, the below problem features of the selected papers have been briefly reviewed in this SLR:

- Type of decision level
- Type of scheduling system
- Type of objective function
- Type of surgery (elective or emergency surgeries)
- Investigation of uncertainty

Table 2 summarizes the selected 28 papers according to the problem features mentioned above. With regard to the decision level, this table shows that the operational decision level (including advance and allocation scheduling problems) has received the most attention amongst these papers. To be more specific, 19 and 18 papers have studied the advance and allocation scheduling problems, respectively; while only five papers have investigated the tactical decision level. According to Table 2, Aringhieri et al. (2015), Spratt and Kozan (2016), Almaneea and Hosny (2018) and Lu et al. (2019) are the only papers that have studied both the tactical and operational decision levels. Among these four papers, Lu et al. (2019) is the only investigation that has studied tactical and operational decision levels, where both advance and allocation scheduling problems are considered. This table also shows that the open scheduling policy has been mostly applied in these papers. The open scheduling system has been used in 19 investigations, while only eight papers have used the block scheduling system. Molina-Pariente et al. (2015) is the only investigation that applied the modified block scheduling system.

Based on Table 2, optimization of resources' utilization is the most common objective function (13 papers); whereas, maximization of revenue is the least common objective function (two papers). Note that the optimization of utilization can refer to the minimization of overtime, minimization of idleness and so on. Now, we classify the decision level of papers based on the scheduling system. Table 2 shows that all papers studying the tactical decision level have applied the block scheduling system (five papers). For the operational decision level, the open scheduling system has been applied mostly (both advance and allocation scheduling problems). The open scheduling system has been used 11 and 14 times for the advance and allocation scheduling problems, respectively; while the block scheduling system has been applied to these problems for only seven and four times, respectively. Table 2 shows that Razmi et al. (2015), Latorre-Núñez et al. (2016) and Belkhamsa et al. (2018) considered both elective and emergency patients. It also shows that the majority of papers did not study the inherent uncertainties of OTPS problems, like surgery duration (only 10 papers included uncertainties).

In the following, we would like to realize whether the decision level or scheduling system impacts the investigation of emergency patients or uncertainties. Figure 4 provides two bar charts to illustrate the percentage of publications studying emergency and uncertainty categorized based on the decision levels and scheduling systems. Note that we have excluded the strategic decision level from Figure 4a because none of the selected papers has considered this decision level.

Table 2. The summary of problems' features for the selected papers

| Paper | Decision level | | | | | Scheduling system | | | Objective function | | | | | | | Patient | | Uncertainty |
|---|---|---|---|---|---|---|---|---|---|---|---|---|---|---|---|---|---|---|
| | S | T | O1 | O2 | O3 | OP | BC | MBC | C | R | U | W | M | NS | OT | E1 | E2 | |
| Marques and Captivo (2015) | | | ✓ | ✓ | | | ✓ | | | | ✓ | | | | ✓ | ✓ | | |
| Molina-Pariente et al. (2015) | | ✓ | | | | | | ✓ | | | | | | | ✓ | ✓ | | |
| Aringhieri et al. (2015) | ✓ | ✓ | | | | | ✓ | | | | | ✓ | | | ✓ | ✓ | | |
| Razmi et al. (2015) | | ✓ | | | | ✓ | | | ✓ | | | | | | | ✓ | ✓ | ✓ |
| Xiang et al. (2015a) | | ✓ | | | | ✓ | | | | | | | ✓ | | | ✓ | | ✓ |
| Xiang et al. (2015b) | | | | ✓ | | ✓ | | | | | | | ✓ | | | ✓ | | |
| Xiang and Li (2015) | | | | ✓ | | ✓ | | | | | ✓ | | | | | ✓ | | |
| Gu et al. (2015) | | | | ✓ | | ✓ | | | | | ✓ | | ✓ | | ✓ | ✓ | | ✓ |
| Beroule et al. (2016) | | | ✓ | | | ✓ | | | | | | | | | ✓ | ✓ | | |
| Spratt and Kozan (2016) | ✓ | ✓ | | | | | ✓ | | | | | | | ✓ | | ✓ | | |
| Latorre-Núñez et al. (2016) | | ✓ | ✓ | | | ✓ | | | | | | | ✓ | | | ✓ | ✓ | ✓ |
| Xiang (2017) | | | ✓ | | | ✓ | | | | | ✓ | | ✓ | | ✓ | ✓ | | ✓ |
| Wu et al. (2018) | | | ✓ | | | ✓ | | | | | | | ✓ | | | ✓ | | |
| Hooshmand et al. (2018) | | | ✓ | ✓ | ✓ | ✓ | | | | | ✓ | | | | | ✓ | | ✓ |
| Nyman and Ripon (2018) | | | ✓ | ✓ | | | ✓ | | | | ✓ | ✓ | | | | ✓ | | |
| Li et al. (2018) | ✓ | | | | | | ✓ | | | | | | | | ✓ | ✓ | | |
| Almaneea and Hosny (2018) | ✓ | ✓ | | | | | ✓ | | ✓ | | ✓ | | | | | ✓ | | |
| Belkhamsa et al. (2018) | | | ✓ | ✓ | | ✓ | | | | | ✓ | ✓ | | | | ✓ | ✓ | |
| Ansarifar et al. (2018) | | | ✓ | ✓ | | | ✓ | | | ✓ | ✓ | | | | ✓ | ✓ | | ✓ |
| Timucin and Birogul (2018) | | | | ✓ | | ✓ | | | | | ✓ | | | | | ✓ | | |
| Nazif (2018) | | | | ✓ | | ✓ | | | | | | | ✓ | | | ✓ | | ✓ |
| Vali-Siar et al. (2018) | | | ✓ | ✓ | | ✓ | | | | | ✓ | ✓ | | | | ✓ | | ✓ |
| Keyhanian et al. (2018) | | | ✓ | ✓ | | ✓ | | | ✓ | | | | | | ✓ | ✓ | | |
| Lin and Chou (2019) | | ✓ | | | | ✓ | | | | | ✓ | ✓ | | | | ✓ | | |
| Lu et al. (2019) | ✓ | ✓ | ✓ | | | | ✓ | | | ✓ | ✓ | | | ✓ | ✓ | ✓ | | |
| Varmazyar et al. (2019) | | ✓ | ✓ | | | ✓ | | | | | | | ✓ | | | ✓ | | ✓ |
| Khalfalli et al. (2019) | | | | ✓ | | | | | | | ✓ | | ✓ | | | ✓ | | |
| Wu et al. (2019) | | ✓ | | ✓ | | ✓ | | | ✓ | | ✓ | | | | | ✓ | | |

S: Strategic; T: Tactical; O1: Operational – advance; O2: Operational – allocation; O3: Online; OP: Open; BC: Block; MBC: Modified block; C: Cost (minimization); R: Revenue (maximization); U: Utilization (optimization); W: Waiting time (minimization); M: Makespan (minimization); NS: Number of surgeries (maximization); OT: Other (optimization); E1: Elective; E2: Emergency.

As shown in Figure 4a, emergency patients have been only studied at the operational decision level amongst 28 reviewed papers. Whereas the uncertainty has been considered at both tactical and operational decision level. This figure shows that studies in the operational decision level have paid more attention to the uncertainty compared to the tactical decision level. Like Figure 4a, Figure 4b shows that researchers have less frequently considered emergency patients. In fact, only three papers (16%) took emergency patients into account, and all of these papers applied the open scheduling system. On the other hand, the figure demonstrates that uncertainty has been investigated under both open and block scheduling systems. Based on Figure 4b, 47% of papers studying the open scheduling system has also investigated uncertainty. This number decreases to 25% for the block scheduling system.

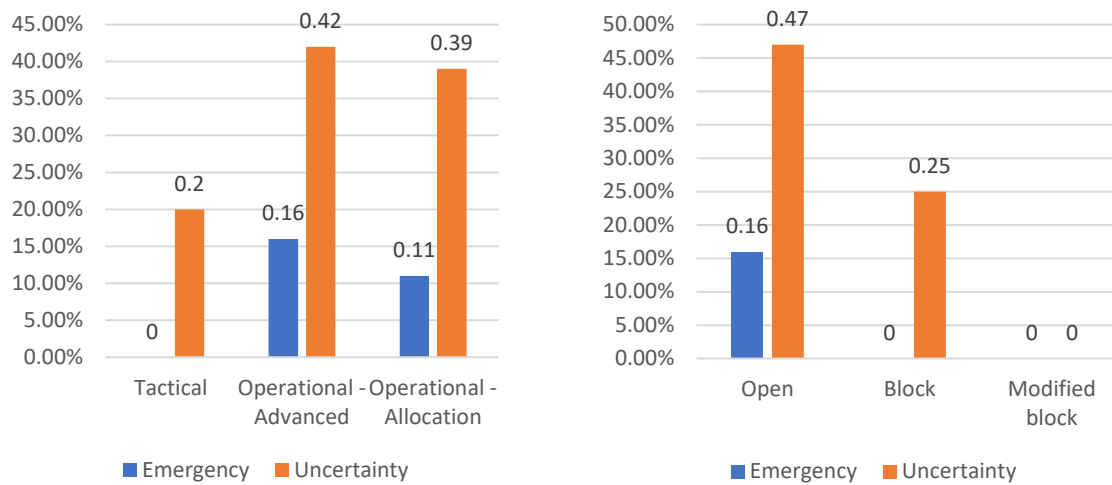

a)  Decision level                    b)  Scheduling system

Figure 4. Percentage of publications studying emergency and uncertainty categorized based on the decision level and scheduling system

## 4. Metaheuristics

The literature consists of a variety of effective solution approaches that have been applied to solving OTPS problems. These solution approaches vary from exact methods to metaheuristics. The choice of the solution approach depends on the size and complexity of the problem. As mentioned earlier, this study focuses on metaheuristics. A metaheuristic is designed to find, generate, or select a heuristic that may provide a sufficiently good solution to an optimization problem, especially with incomplete or imperfect information or limited computation capacity (Varmazyar et al., 2019). Metaheuristics may make few assumptions about the optimization problem being solved, and so they may be usable for a variety of problems. Compared to optimization algorithms and iterative methods, metaheuristics do not guarantee that a globally optimal solution can be found. Many metaheuristics implement some form of stochastic optimization, so that the solution found is dependent on the set of random variables generated.

Metaheuristics applied to OTPS problems have been reviewed in the rest of this paper. For this purpose, we first aim to categorize our set of papers based on the type of metaheuristic algorithms. Figure 5 represents the frequency of the application of metaheuristics. According to this figure, Genetic Algorithm (GA) and Ant Colony Optimization (ACO) have been used most frequently (11 and 6 times, respectively). On the other hand, each of the Bees, Deferential Evolution (DE), Random Extraction-Insertion (REI) and Iterative Local Search (ILS) algorithms have been merely used for once. In addition, Particle Swarm Optimization (PSO), Simulated Annealing (SA), and Tabu Search (TS) are three other algorithms that have been applied to OTPS problems for more than once. The sum of the frequencies in Figure 5 equals 31; while we have reviewed 28 papers. This is because three papers have reported the application of two types of metaheuristics.

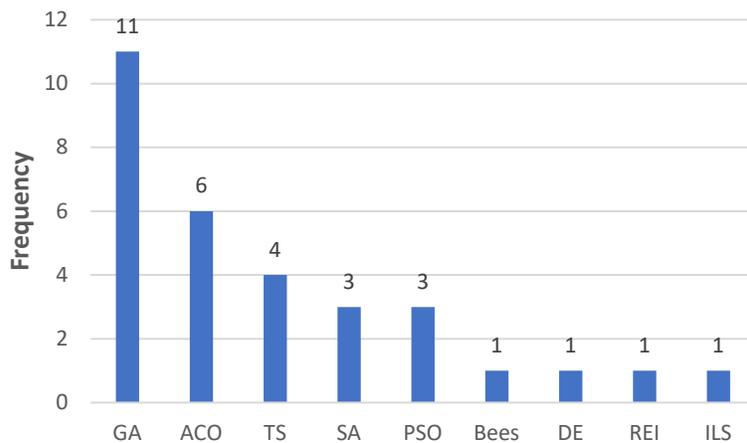

Figure 5. Frequency of the use of metaheuristics

In the following, we would like to discuss the features of the applied metaheuristics in detail. Metaheuristics have different structures and procedures. For example, they could vary in terms of their solution representations, termination condition, initial solution generation and so forth. For this reason, we need to define a set of features so as to evaluate different metaheuristics based on them. These features are as follows:

- *Solution structure*: The structure of the solution(s) in a metaheuristic can vary. One may use a one-part solution structure, another one would use a multiple parts solution structure.

- *Solution shape*: A solution can be represented in the form of an array of numbers or a matrix of numbers.

- *Solution value*: Metaheuristics can use different types of values to define solutions. They may use real, integer, or binary numbers in their solution representations.

- *Number of solutions*: Metaheuristics may be a single solution algorithm (having only one solution in each iteration) or they would be a population-based algorithm (having a population of solutions in each iteration).

- *Initial solution*: Metaheuristics need an initial solution(s) to start the process of optimization. Generally, initial solutions are generated randomly or based on a constructive algorithm.

- *Replacement*: It is a process through which a new solution(s) is replaced with the incumbent solution(s). In this process, the new solution may be compared with its parent or the entire population of incumbent solutions.

- *Parent selection*: In some metaheuristics, new solutions are generated from a number of selected parents. We recognized different methods for the selection of parents, like random, tournament and roulette wheel.

- *Feasibility of solution*: There are usually two types of metaheuristics: (1) algorithms that guarantee the feasibility of solutions, and (2) algorithms that do not guarantee the feasibility of solutions. For the latter group, different methods can be incorporated to guarantee the feasibility of solutions, such as penalty function, repair function, and elimination of infeasible solutions.

- *Termination criterion*: This is a criterion to determine when the search process of a metaheuristic should be terminated. We have recognized five categories of termination criteria: (1) maximum number of iterations, (2) number of consecutive iterations without improvement, (3) finding a solution with a desired objective function value, (4) minimum temperature, (5) other.

We will use the abovementioned features to analyze the set of 28 selected papers. Table 3 summarizes the papers according to the abovementioned features of solution approaches. Based on this table, one-part and multi-part solution structures have gained almost equal attention. Eleven metaheuristics have used one-part solution structures, and 13 metaheuristics have applied multi-part solution structures. Table 3 shows that seven metaheuristics have not reported their solution structure. This table also demonstrates that most of the metaheuristics have used an array of numbers in their solution representation (22 cases); only one metaheuristic has benefited from a matrix-based shape solution. Eight metaheuristics have not stated their solution shape. In terms of the solution value, most metaheuristics have used integer numbers for the representation of solutions (19 cases). Only one metaheuristic has benefited from binary numbers.

Like solution shapes, eight metaheuristics have not indicated their solution values. Based on Table 3, 22 algorithms were population-based metaheuristics and nine algorithms were single solution metaheuristics. The population-based algorithms include GA, Bees, ACO, PSO and DE, and the single solution algorithm includes SA, TA, ILS and REI. The application of random (ten cases) and constructive (11 cases) procedures for the generation of initial solutions has gained equal attention from researchers. Regarding the replacement of new solutions, Table 3 reports that the majority of metaheuristics have not indicated their methods. However, 11 algorithms have a replacement strategy that compares new solutions with the population of incumbent solutions, and only two algorithms compare new solutions with their parents.

Table 3. The summary of solution approaches' features for the selected papers

| Paper | Solution approach | Solution representation | | | | | | | Number of solution | | Initial solution | Parent selection | Replacement | | Feasibility of solutions | | | | Termination criterion | Test instances | | | Benchmark solution approach |
|---|---|---|---|---|---|---|---|---|---|---|---|---|---|---|---|---|---|---|---|---|---|---|---|
| | | SST | | SSP | | SVL | | | | | | | | | | | | | | | | | |
| | | OP | MP | AR | MA | BN | IT | RL | SN | Pop | | | Cpop | Cpar | GR | PF | RF | EL | | RN | BM | RC | |
| Marques and Captivo (2015) | GA | ✓ | | ✓ | | | ✓ | | | ✓ | CN | TN | ✓ | | ✓ | | | | MI | | | ✓ | |
| Molina-Pariente et al. (2015) | REI | ✓ | | ✓ | | | ✓ | | ✓ | | CN | NS | | | ✓ | | | | MI | | ✓ | | H/M |
| Aringhieri et al. (2015) | TS | | | | | | | | ✓ | | CN | NS | | | ✓ | | | | NI | | | ✓ | MP |
| Razmi et al. (2015) | DE | | | | | | | ✓ | | ✓ | RN | RN | | ✓ | ✓ | | | | DV | | | ✓ | |
| Xiang et al. (2015a) | ACO | | ✓ | ✓ | | | ✓ | | | ✓ | | NS | | | ✓ | | | | MI | ✓ | | | SM |
| Xiang et al. (2015b) | ACO | | ✓ | ✓ | | | ✓ | | | ✓ | | NS | | | ✓ | | | | MI | | ✓ | ✓ | H/M |
| Xiang and Li (2015) | ACO | | ✓ | ✓ | | | ✓ | | | ✓ | | NS | | | ✓ | | | | MI | | | ✓ | |
| Gu et al. (2015) | ACO | | ✓ | ✓ | | | ✓ | | | ✓ | | NS | | | ✓ | | | | MI | | ✓ | ✓ | SM |
| Beroule et al. (2016) | PSO | ✓ | | ✓ | | | | ✓ | | ✓ | RN | NS | | | ✓ | | | | MI | | | ✓ | |
| Spratt and Kozan (2016) | SA | | | | | | | | ✓ | | CN | NS | | | | | | ✓ | MI | | | ✓ | |
| Latorre-Núñez et al. (2016) | GA | ✓ | | ✓ | | | | | | ✓ | RN | NS | ✓ | | ✓ | | | | MI | ✓ | | | |
| Xiang (2017) | ACO | | ✓ | ✓ | | | ✓ | | | ✓ | | NS | | | ✓ | | | | MI | | ✓ | ✓ | SM |
| Wu et al. (2018) | GA | ✓ | | ✓ | | | ✓ | | | ✓ | RN | RW | ✓ | | | | ✓ | | MI | | | ✓ | |
| Hooshmand et al. (2018) | GA | ✓ | | ✓ | | | ✓ | | | ✓ | CN | TN | ✓ | | | ✓ | | | MI | ✓ | | | MP |
| Nyman and Ripon (2018) | GA | | ✓ | ✓ | | | ✓ | | | ✓ | RN | TN | | | ✓ | | | | MI | | | ✓ | |
| Li et al. (2018) | TS | ✓ | | ✓ | | | ✓ | | ✓ | | CN | | | | ✓ | | | | MI | ✓ | ✓ | | MP |
| Almaneea and Hosny (2018) | Bees | | | | | | | | | ✓ | RN | | ✓ | | | | | | NI | | ✓ | | |
| Belkhamsa et al. (2018) | GA | ✓ | | ✓ | | | ✓ | | | ✓ | RN | | ✓ | | ✓ | | | | MI | | ✓ | | H/M |
| | ILS | ✓ | | ✓ | | | ✓ | | ✓ | | CN | NS | | ✓ | ✓ | | | | MI | | ✓ | | H/M |
| Ansarifar et al. (2018) | GA | | ✓ | ✓ | | | | ✓ | | ✓ | RN | RN | ✓ | | | | | | | | | ✓ | SM |
| | PSO | | ✓ | ✓ | | | | ✓ | | ✓ | | NS | ✓ | | | | | | | | | ✓ | SM |
| Timucin and Birogul (2018) | GA | ✓ | | | ✓ | | ✓ | | | ✓ | | RW | ✓ | | | | ✓ | | DV | | | | |
| Nazif (2018) | ACO | | ✓ | ✓ | | | ✓ | | | ✓ | | NS | ✓ | | | | | | MI | | | ✓ | |
| Vali-Siar et al. (2018) | GA | | ✓ | ✓ | | | ✓ | | | ✓ | CN | RW | ✓ | | | | | | MI | | ✓ | ✓ | H/M |
| Keyhanian et al. (2018) | SA | | ✓ | ✓ | | | ✓ | | ✓ | | RN | | | | ✓ | | | | MT | ✓ | | | |
| Lin and Chou (2019) | GA | ✓ | | ✓ | | | ✓ | | | ✓ | RN | | | | | ✓ | | | OT | ✓ | | | MP |
| Lu et al. (2019) | GA | | ✓ | ✓ | | ✓ | | | | ✓ | CN | TN | ✓ | | | | ✓ | | MI | | | ✓ | MP |
| Varmazyar et al. (2019) | TS | | | | | | | | ✓ | | | | | | ✓ | | | | NI | ✓ | | | MP |
| | SA | | | | | | | | ✓ | | | | | | ✓ | | | | MT | ✓ | | | |
| Khalfalli et al. (2019) | TS | | | | | | | | ✓ | | CN | | | | ✓ | | | | MI | | | ✓ | |
| Wu et al. (2019) | PSO | | ✓ | ✓ | | | ✓ | | | ✓ | RN | NS | | | | | ✓ | | MI | | | ✓ | H/M |

SST: Solution structure; OP: One-part; MP: Multi-part; SSP: Solution shape; AR: Array; MA: Matrix; SVL: Solution value; BN: Binary; IT: Integer; RL: Real; SN: Single solution; Pop: Population-based; RN: Random; CN: Constructive; NW: No selection; RW: Roulette wheel; TN: Tournament; Cpop: Comparison with population; Cpar: Comparison with parent; GR: Guaranteed; PF: Penalty function; RF: Repair function; EL: Elimination; MI: Maximum number of iteration; NI: Number of consecutive iterations without improvement; DV: Desired value; MT: Minimum temperature; OT: Other; BM: Benchmark; RC: Real case; MP: Mathematical programming; SM: Simulation modeling; H/M: Heuristic or Metaheuristic.

If a problem studies more than one objective function, the optimization process would not be as straightforward as a problem with one objective function. In our set of papers, 19 metaheuristics have been applied to single objective function problems. For multi-objective function problems, weighted sum or non-dominated sorting techniques have been used for seven and five times, respectively. Finally, test instances are essential for the evaluation of metaheuristics. Considering the set of selected papers, we have found three types of test instances: (1) randomly generated, (2) benchmark, and (3) real-world instances. Real-world instances have been used most frequently (18 times), and benchmark tests were used least frequently. Randomly generated and benchmark instances have been also used eight and five times, respectively.

We have studied for other characteristics of metaheuristics in the following, including parent selection, feasibility of solution, termination criterion and benchmark solution approaches. Figure 6 illustrates the frequency of different methods used for each of these characteristics. With respect to parent selection, Figure 6a shows that the majority of metaheuristics did not require a procedure for the parent selection (14 cases). This is because these metaheuristics can be single solution algorithms, or their entire population will be selected as parents. Amongst those metaheuristics requiring a parent selection method, the tournament method was used most frequently (four times). The roulette wheel and random methods have been also applied to metaheuristics (three times and twice, respectively). According to Figure 6b, most metaheuristics guaranteed the feasibility of solutions (19 cases), which emphasizes the importance of the feasibility of solutions from the perspective of researchers. Metaheuristics have also used penalty function (twice), repair function (four times) and elimination of infeasible solutions (once) to ensure the feasibility of solutions provided by metaheuristics.

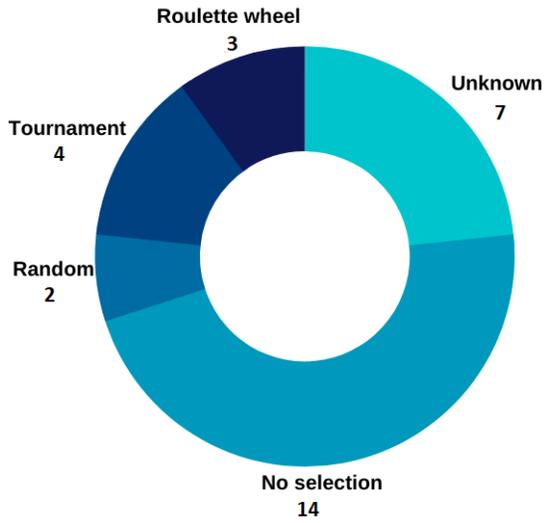

a) Parent selection

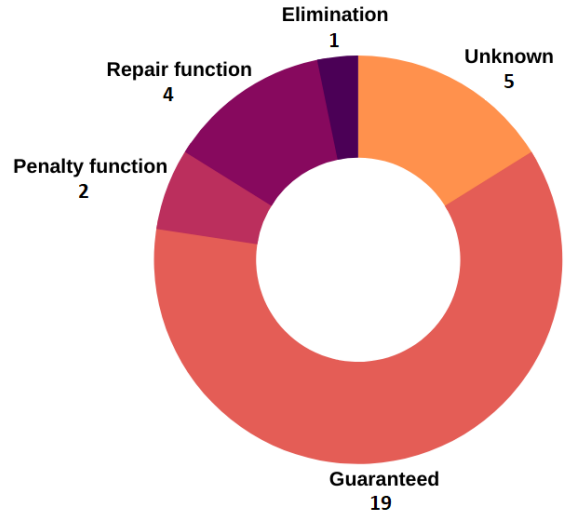

b) Feasibility of solution

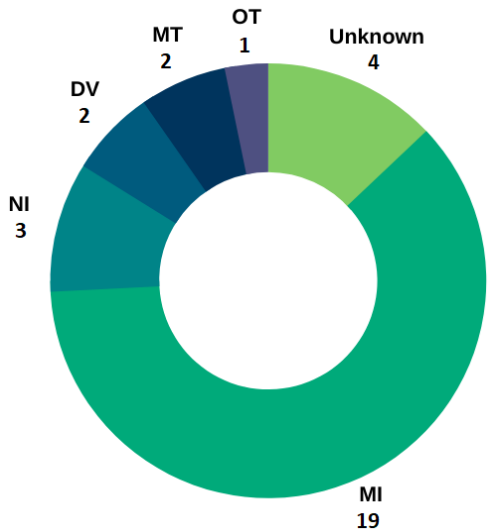

c) Termination criterion

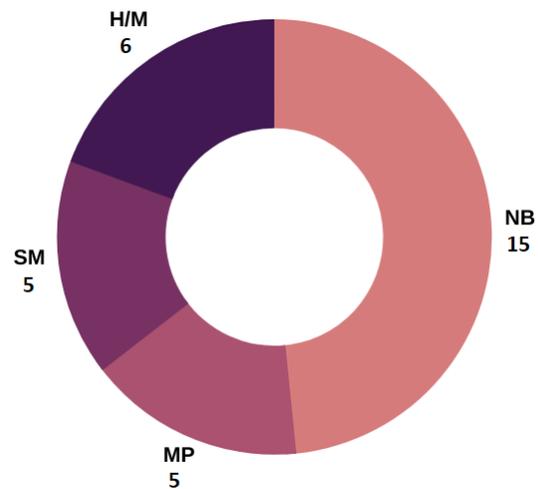

d) Benchmark solution approach

MI: Maximum number of iteration; NI: Number of consecutive iterations without improvement; DV: Desired value; MT: Minimum temperature; OT: Other.

NB: No benchmark; MP: Mathematical programming; SM: Simulation modeling; H/M: Heuristics or Metaheuristics

Figure 6. Parent selection, feasibility of solution, termination criterion, and benchmark solution approaches of metaheuristics

Figure 6c also reports on different methods used as the termination criterion for metaheuristics. Based on this figure, the maximum number of iterations was the most widely used method of the termination criterion (19 cases). The number of consecutive iterations without improvement, desired value and minimum temperature are three termination criterion methods used in our set of papers. Finally, Figure 6d indicates whether a benchmark solution approach has been applied to compare with metaheuristics. The figure shows that the majority of metaheuristics were compared with a benchmark solution approach. In fact, five metaheuristics were compared with mathematical programming, five metaheuristics were compared with simulation modeling, and six metaheuristics were compared with heuristics/metaheuristics. However, 15 metaheuristics have not been compared with any other benchmark solution approaches.

To have a more detailed analysis, Figure 7 categorizes the use of metaheuristics based on different types of decision levels and scheduling system. As shown in Figure 7a, only four metaheuristics have been deployed to investigate the tactical decision making, including GA, Bees, SA and TS. Nevertheless, all metaheuristics shown in Figure 7 have been used to study the operational decision-making level. Figure 7a also demonstrates that the tactical decision level was mostly studied by single solution metaheuristics (one case with SA and two cases with TS). In contrast, the operational decision level was mostly investigated by population-based metaheuristics (in total, 19 cases with GA, five cases with ACO, and five cases with PSO). Figure 7b does not reveal any meaningful difference in the application of metaheuristics (except for ACO) for different scheduling systems. For example, both SA and TS have been used for the investigation of open and block scheduling systems twice. As the mere exception, ACO is only applied to the open scheduling system for six times.

Metaheuristics can be investigated with regard to the frequency of their applications categorized based on the consideration of emergency patients and uncertainty. Analysis of extracted data shows that the investigation of emergency patients or uncertainty was mostly done through single solution metaheuristics. Emergency patients were only studied using single solution metaheuristics. Both single solution and population-based algorithms have been applied for the investigation of uncertainty. Respecting single solution algorithms, REI, SA and TS have been developed once, twice and twice, respectively. For the population-based algorithm, six applications have been reported for both GA and ACO in total.

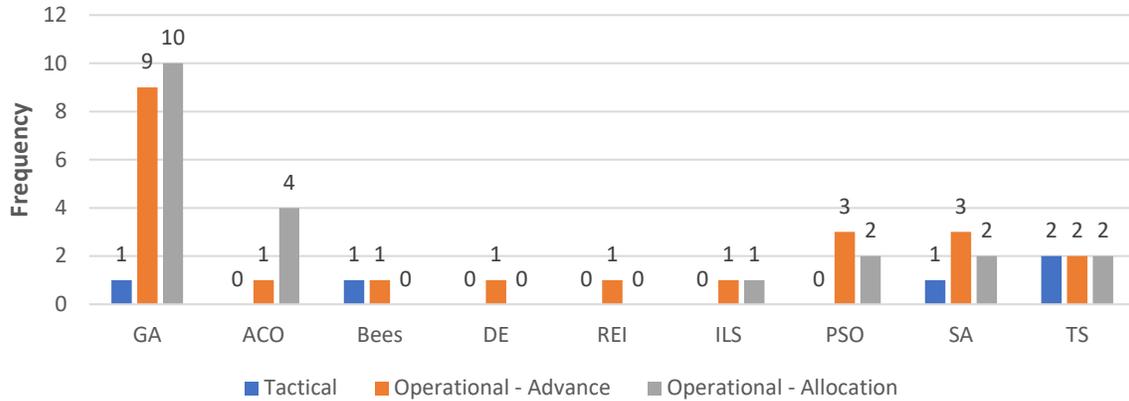

a) Decision level

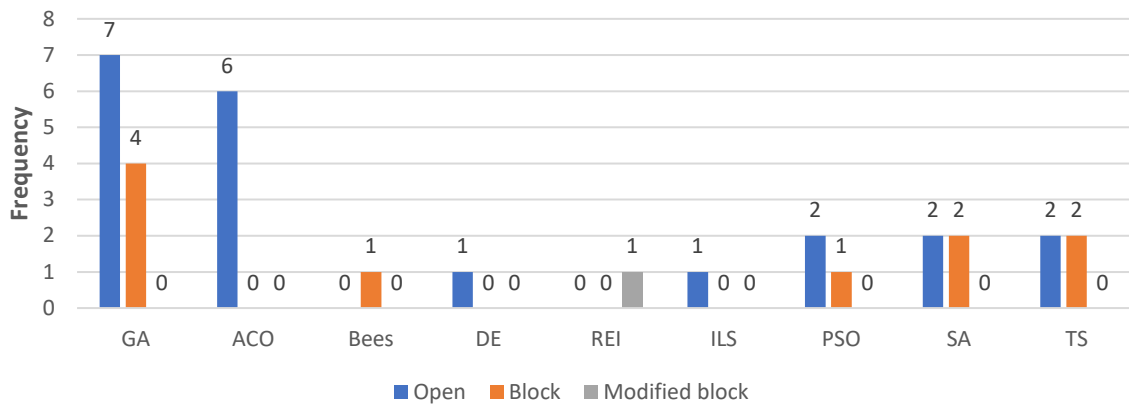

b) Scheduling system

Figure 7. Frequency of the use of metaheuristics categorized based on the decision level and scheduling system

Up to this point, we showed that most metaheuristics suffer from a lack of clarity in introducing their design procedures. To elaborate more on this issue, Figure 8 illustrates the lack of information provided for metaheuristics categorized based on their features and types. Considering Figure 8a, we have found that most metaheuristics are suffering from the lack of clarity for at least one of their features. The replacement method is the most poorly explained feature (18 cases). While the number of solutions is the only feature explained in all applications of metaheuristics. Thus, it can be argued that metaheuristics are not explained well in previous research. It means that practitioners or other researchers may not be able to reproduce the same metaheuristics for the use of real-world situations or future research. That is why we would like to call future research to make more effort in order to clarify every characteristic of their solution approaches. Figure 8b reports the average number of features has not been explained for metaheuristics. According to this figure, each of TS, REI and Bees algorithms have not introduced about four of their features on average. This number decreases to 0.91 of features for GA and SA on average. Finally, Figure 8b shows that ILS is the only metaheuristic that does not suffer from a lack of clarity.

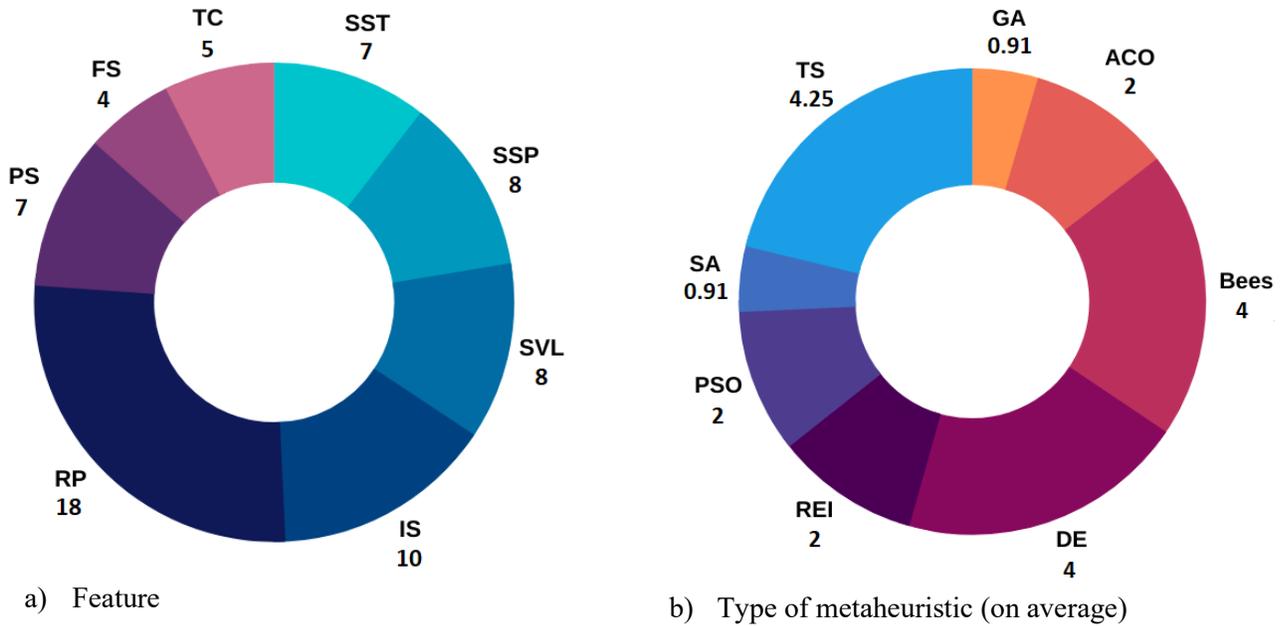

a) Feature

b) Type of metaheuristic (on average)

SST: Solution structure; SSP: Solution shape; SLV: Solution value; IS: Initial solution; RP: Replacement; PS: Parent selection; FS: Feasibility of solution; TC: Termination criterion.

Figure 8. Frequency of lack of provided information categorized based on features and types of metaheuristic

Figure 6d revealed that 15 metaheuristics were not compared with a benchmark solution approach. Under this circumstance, they should be evaluated using benchmark or real-world test instances. Otherwise, their performance may be under question. For this reason, we have first identified metaheuristics that are not compared with a benchmark solution approach. Then, we have categorized them based on their test instances. We found that most of the metaheuristics are only compared with randomly generated test instances (11 metaheuristics). At the same time, only six metaheuristics were evaluated using either benchmark or real-world test instances. Therefore, we would like to call future research for the deployment of benchmark solution approaches or benchmark or real-world test instances, which can increase their credibility.

Figure 6b demonstrated a strong interest of previous researchers in metaheuristics that guarantee the feasibility of solutions. For this reason, we would like to have a more in-depth analysis of the features of such metaheuristics. Table 4 compares the metaheuristics that have guaranteed the feasibility of solutions with respect to: (1) solution structure, (2) solution shape, (3) solution value, (4) number of solutions, and (5) initial solution. According to this table, most of these algorithms have used a one-part solution representation shaped with an array of integer numbers. With regard to the number of solutions, Table 4 shows that eight applications of single solution metaheuristics have been reported for algorithms that have guaranteed the feasibility of solutions. This is while we found nine applications of single solution metaheuristics in total. It reveals the interests of researchers in the application of single solution metaheuristics to ensure the feasibility of solutions.

Table 4. Features of metaheuristics that guaranteed the feasibility of the solution

| Feature | Detail |
| --- | --- |
| Solution structure | Unknown: 4 metaheuristics<br>One-part: 7 metaheuristics<br>Multi-part: 8 metaheuristics |
| Solution shape | Unknown: 5 metaheuristics<br>Array: 14 metaheuristics<br>Matrix: 0 metaheuristic |
| Solution values | Unknown: 5 metaheuristics<br>Real: 1 metaheuristic<br>Integer: 13 metaheuristics<br>Binary: 0 metaheuristic |
| Number of solutions | Single solution: 8 metaheuristics<br>Population-based: 11 metaheuristics |
| Initial solution | Unknown: 8 metaheuristics<br>Random: 5 metaheuristics<br>Constructive: 6 metaheuristics |

## 5. Conclusions, limitations and future research directions

In this paper, we reviewed solution approaches developed metaheuristic algorithms for OTPS problems between 2015 and 2020. First, we defined a search query; then, by conducting an automatic search through four major databases, 294 papers were found. After the elimination of duplicates, application of the exclusion and inclusion criteria and snowballing method, 42 papers remained for the next step, from which 14 were eliminated due to the application of the quality assessment. Finally, 28 papers were selected for the review. We studied the features of the problems considered in these papers and compared them. The features of metaheuristics were also examined and reported. Through these examinations we found some key points and gaps in the literature summarized as follows:

- Most of the solution approaches proposed in the literature are deprived of a mature diversity mechanism, while benefiting such a mechanism is somewhat necessary to avoid premature convergence.

- Most of the solution approaches have been developed for problems in only one of the decision levels. ACO, for instance, have been proposed only for allocation scheduling problem. Therefore, developing ACO algorithm for an advance scheduling problem or multiple decision-levelsat the same time can be interesting for future research. Like ACO, the situation for other solution approaches is the same.

- Having applied different scheduling systems can result in different features of solution approaches; most of the papers applied open scheduling system. Only REI has been proposed for the modified block scheduling systems. Therefore, this area has considerable potential for future research. Moreover, someone can notice some of the proposed metaheuristic approaches such as ACO and DE have not developed for block scheduling yet.

- Other types of metaheuristics (Bat, Imperialist competitive, Artificial algae, Fireworks, Pigeon-inspired optimization, Brain storm optimization and Earthworm optimization algorithms and so on) were not applied to OTPS problems within the review period of this paper. Thus, the application of such algorithms may be a direction for future research.

- In our review, several papers did not deploy a benchmark solution approach and benchmark test instance. Under this condition, the performance of a solution approach can be questioned. Thus, we would like to call future research for the deployment of benchmark solution approaches or benchmark or real-world test instances, which can increase the credibility of their results.

- We have found that most metaheuristics are suffering from the lack of clarity for at least one of their features. In other words, it can be argued that metaheuristics are not explained well in previous research. It means that practitioners or other researchers may not be able to reproduce the same metaheuristics for the use of real-world situations or future research. That is why we would like to call future research to make more effort in order to clarify every characteristic of their solution approaches.

- Finally, we would like to call for the development of a number of state-of-the-art solution approaches for a unique problem (as a benchmark problem) and comparison of their performances. This comparison may give a better idea of how these algorithms perform in comparison to each other.

An important limitation of this research relates to the internal validity of the results. This is because of the elimination of "Surgery planning", "Surgery scheduling" and "Surgery sequencing" from the search query, which could have led to the loss of some relevant papers. To mitigate the impact of this threat, we have, however, applied a manual search method (snowballing method) after the implementation of the automated search method. Another import threat is that this is a research view based on the scientific literature; reality in practice might be different, and other unpublished metaheuristic methods may exist (and future work could involve an industrial survey to gather that information, for example). Finally, one may argue that the elimination of papers published before 2015 (due to one of the exclusion criteria) has restricted the number of papers included in this review, and 28 papers are not enough for a review paper. However, we provided evidence of limited applications of metaheuristics in the literature.